\documentclass{article}
\usepackage{spconf,amsmath,graphicx}
\usepackage{bbding}

\title{A Video Anomaly Detection Framework based on Appearance-Motion Semantics Representation Consistency}

\name{Xiangyu Huang$^{1}$, Caidan Zhao$^{1*}$ \thanks{*Corresponding Author. This work was supported in part by the National Natural Science Foundation of China under Grant No. 61971368, No. U20A20162 and No. 61731012, and in part by the Natural Science Foundation of Fujian Province of China No. 2019J01003.}, and Zhiqiang Wu$^{2}$}

\address{$^{1}$ School of Informatics, Xiamen University \\
$^{2}$ PKU-Wuhan Institute for Artificial Intelligence}
\begin{document}
%
\maketitle
\begin{abstract}
Video anomaly detection is an essential but challenging task. The prevalent methods mainly investigate the reconstruction difference between normal and abnormal patterns but ignore the semantics consistency between appearance and motion information of behavior patterns, making the results highly dependent on the local context of frame sequences and lacking the understanding of behavior semantics. To address this issue, we propose a framework of Appearance-Motion Semantics Representation Consistency that uses the gap of appearance and motion semantic representation consistency between normal and abnormal data. The two-stream structure is designed to encode the appearance and motion information representation of normal samples, and a novel consistency loss is proposed to enhance the consistency of feature semantics so that anomalies with low consistency can be identified. Moreover, the lower consistency features of anomalies can be used to deteriorate the quality of the predicted frame, which makes anomalies easier to spot. Experimental results demonstrate the effectiveness of the proposed method.
\end{abstract}
\section{Introduction}
\label{sec:intro}

Video anomaly detection (VAD) refers to identifying events that do not conform to expected behavior \cite{chandola2009anomaly}, which is of great practical value in public safety scenarios. In addition to much effort devoted in \cite{luo2017revisit, liu2018future}, VAD remains an extremely challenging task due to the rarity and ambiguity of anomalies \cite{chandola2009anomaly}. It is infeasible to collect balanced normal and abnormal samples and tackle this task with a supervised binary classification model. Therefore, a typical solution to VAD is often formulated as an unsupervised learning problem, where the goal is to train a model by using only normal data to mine regular patterns. The events that do not conform to this model are viewed as anomalies. 
\begin{figure}[ht]
	\centering
	\includegraphics[width=9cm]{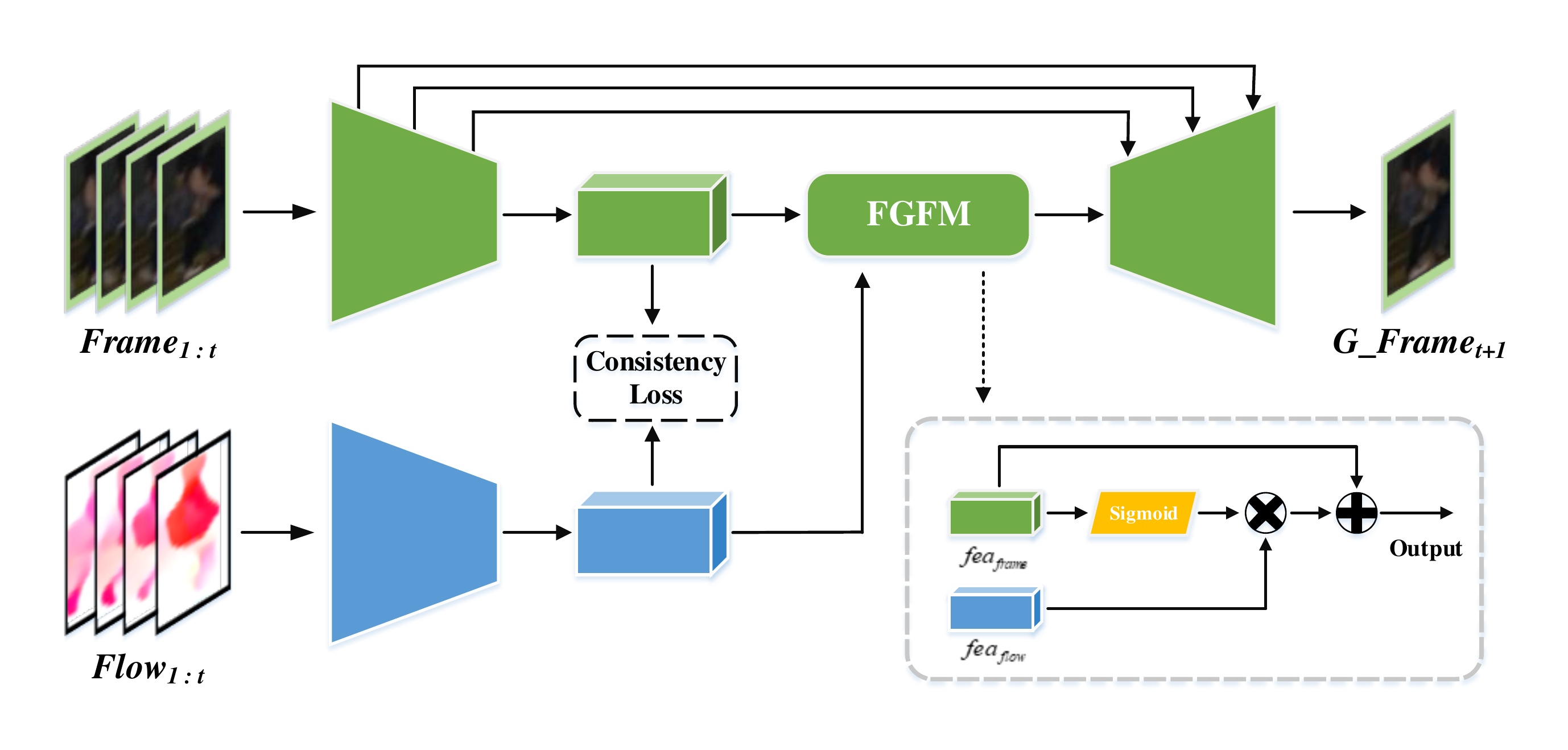}
	\caption{Overview of the proposed AMSRC-Net.}
	\label{p1}
\end{figure}

Existing prevalent VAD methods almost follow a reconstruction or future frame prediction mode. Reconstruction-based methods \cite{hasan2016learning, luo2017remembering, gong2019memorizing, park2020learning} usually train AutoEncoders (AEs) on normal data and expect abnormal data to incur larger reconstruction errors at test time, making abnormal data detectable from normal ones. Prediction-based method \cite{liu2018future} uses the temporal characteristics of video frames to predict the next frame based on a given sequence of previous frames, then uses the prediction errors for anomaly measuring. However, such approaches are highly dependent on the local context of frame sequence, and some studies \cite{gong2019memorizing, liu2021hybrid} have shown that AEs trained only on normal data sometimes reconstruct anomalies well, which leads to poor performance. So some researchers \cite{xu2017detecting, yan2018abnormal, vu2019robust} try to utilize the motion information of activities, which contains a lot of semantics representing behavioral properties, to achieve good performance for VAD. Nevertheless, these methods only combine the information of appearance and motion to detect anomalies in the test phase and do not jointly model the two types of information in the same space during the training phase \cite{xu2017detecting, yan2018abnormal, vu2019robust, nguyen2019anomaly, ye2019anopcn}, which makes it difficult to capture the correlation between the two modalities for VAD. Recently, these state-of-the-art (SOTA) methods \cite{liu2021hybrid, cai2021appearance} are proposed to model the relationship of two modalities and can detect anomalies in most cases, but the results are still far from expectations due to the unstable relationship modeling method. For example, Cai et al. \cite{cai2021appearance} use a memory network to store the relationship of two modalities so that unsuitable-sized memory modules easily limit the network’s performance. Furthermore, Liu et al. \cite{liu2021hybrid} use a hybrid framework in a combination of flow reconstruction and frame prediction, but the result is highly dependent on the quality of flow reconstruction in the previous phase, which makes it difficult to train a stable model. Moreover, the correlations modeled by these methods are essentially designed for recovering the pixel information but still lack the understanding of behavior semantics.

In this paper, we take a step forward in making full use of the multi-modal knowledge from normal events to detect anomalies via a simple yet novel network based on Appearance-Motion Semantics Representation Consistency, termed AMSRC-Net. Specifically, inspired by SOTA methods that use multiple modalities \cite{liu2021hybrid, yan2018abnormal, nguyen2019anomaly, cai2021appearance}, we first extract the representative features of appearance and motion in normal events by a universal two-stream encoder. Unlike previous works, we observe the consistency between the features of two modalities and propose a novel consistency loss to model the semantics consistency in the feature space explicitly. Then, the proposed network generates lower consistent features for abnormal samples, which typically reflect irregular behavior semantics and can be used to detect anomalies. Moreover, we design a simple flow-guided fusion module, which utilizes the above feature semantics consistency gap to augment the prediction quality gap. Extensive experiments on three public VAD datasets show that our proposed AMSRC-Net achieves better performance than SOTA methods.

\section{Proposed Method}
As shown in Figure \ref{p1}, our proposed AMSRC-Net consists of three parts:  A two-stream encoder, a decoder, and a flow-guided fusion module (FGFM). We first input the previous video frame image and its optical flow clip into the two-stream encoder to get the appearance and motion’s feature representations. Then the proposed consistency loss is used to enhance further the consistency of the feature semantics between appearance and motion information in normal samples. Next, two consistent modality features are put into the flow-guided fusion module. Finally, feeding the fused feature into the decoder to predict the future frame image. The detailed network architecture of AMSRC-Net is shown in Figure 2 and all the components are presented in the following subsections in detail.

\noindent
\textbf{Two-stream Encoder and Decoder.}
The two-stream encoder extracts feature representations from input video frame images and the corresponding optical flows. Due to the consistency constraints, the extracted features’ semantics are highly similar, representing the foreground behavior properties in the surveillance video. Then the decoder is trained to generate the next frame by taking the aggregated feature formed by fusing the extracted features from the previous step. The aggregated feature may lack low-level information, such as backgrounds, textures, and so on. To solve this problem, we add a UNet-like skip connection structure \cite{ronneberger2015u} between the frame stream encoder and decoder to preserve these low-level features irrelevant to behavior for predicting the high-quality future frame.

\begin{figure*}[ht]
	\centering
	\includegraphics[width=14cm]{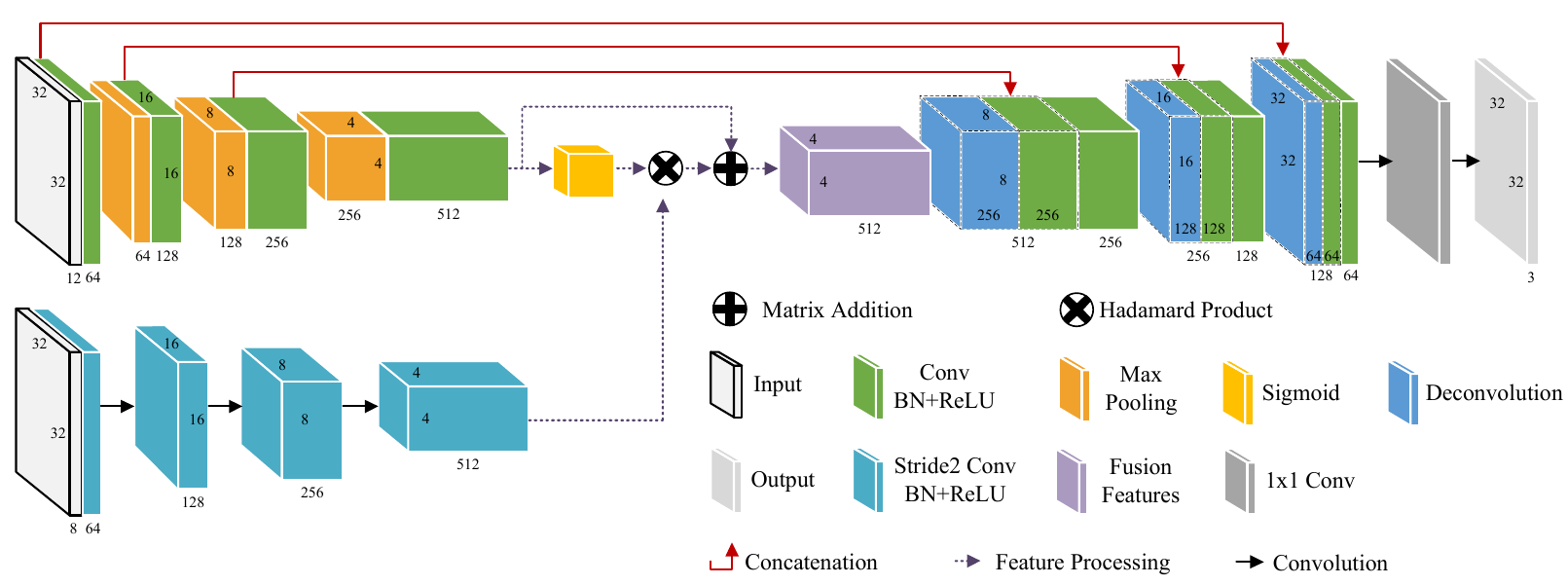}
	\caption{Detailed network architecture of AMSRC-Net.}
	\label{p2}
\end{figure*}

\noindent
\textbf{FGFM. }
Since ReLU activation is adopted at the end of the two-stream encoder, many feature representations have zero value in the output features. During the training for the semantics consistency of two-stream features, we observe that the distribution of two-stream features over non-zero feature representations is highly consistent. In contrast, the lower consistency of appearance-motion features generated by abnormal data reflects a larger difference in the distribution of two-stream features over non-zero feature representations. In order to utilize the above feature representation gap to improve the VAD performance, we design a simple flow-guided fusion module to enlarge the prediction error gap between normal and abnormal samples. Given appearance features $fea_{frame}$ and motion features $fea_{flow}$, we use Hadamard product between the activation of $fea_{frame}$ and $fea_{flow}$ without the linear projection and residual operation to produce the fused feature $fea_{fused}$, which is used for prediction:

\begin{equation}
fea_{fused}= fea_{frame} \oplus \left(\sigma\left(fea_{frame}\right) \otimes fea_{flow}\right)
\end{equation}
where $\sigma$ denotes Sigmoid function, $\oplus$ and $\otimes$ denote Matrix Addition and Hadamard product, respectively. There is a gap in the fused feature representation between normal and abnormal data, and only the fused feature of normal data is trained to generate a high-quality future frame. With the increase of the gap in the fused feature representation during training, the gap in the quality of the predicted frame is also enlarged.

\noindent
\textbf{Loss Function.}
We follow the previous VAD work based on prediction \cite{liu2018future}, using intensity and gradient difference to make the prediction close to its ground truth. The intensity loss guarantees the similarity of pixels between the prediction and its ground truth, and the gradient loss can sharpen the predicted images. We minimize the $\ell_{2}$ distance between the predicted frame $\hat{x}$ and its ground truth $x$ as follows:
\begin{equation}
	L_{i n t}=\left\|\hat{x}-x\right\|_{2}^{2}
\end{equation}

The gradient loss is defined as follows:
\begin{equation}
	\begin{aligned}
		L_{g d}=\sum_{i, j}&\left\|\left|\hat{x}_{i, j}-\hat{x}_{i-1, j}\right|-\left|x_{i, j}-x_{i-1, j}\right|\right\|_{1}\\
		+&\left\|\left|\hat{x}_{i, j}-\hat{x}_{i, j-1}\right|-\left|x_{i, j}-x_{i, j-1}\right|\right\|_{1}
	\end{aligned}
\end{equation}
where $i$, $j$ denote the spatial index of a video frame.

In order to model the appearance and motion semantic representation consistency of normal samples, we minimize the cosine distance between the appearance and motion features of normal samples encoded by the two-steam encoder. The proposed consistency loss is defined as follows:
\begin{equation}
	L_{sim}=1-\frac{\langle{fea_{frame}}, fea_{flow}\rangle}{\|fea_{frame}\left\|_{2}\right\|fea_{flow}\|_{2}}
\end{equation}

Then, the overall loss $L$ takes the form as follows:
\begin{equation}
	L=\lambda_{int} L_{i n t}+\lambda_{g d} L_{g d}+ \lambda_{sim} L_{sim}+\lambda_{model} \left\|W\right\|_{2}^{2}
\end{equation}
where $\lambda_{int}$, $\lambda_{gd}$, and $\lambda_{sim}$ are balancing hyper-parameters, $W$ is the parameter of the model, and $\lambda_{model}$ is a regularization hyper-parameter that controls the model complexity. 

\noindent
\textbf{Anomaly Detection.}
Our anomaly score is composed of two parts during the testing phase: the inconsistency between appearance and motion feature $S_{f}=1-\frac{\langle{fea_{frame}}, fea_{flow}\rangle}{\|fea_{frame}\left\|_{2}\right\|fea_{flow}\|_{2}}$ and the future frame prediction error $S_{p}=\left\|\hat{x}-x\right\|_{2}^{2}$. Then, we get the final anomaly score by fusing the two parts using a weighted sum strategy as follows:
\begin{equation} \label{e5}
	\mathrm{S}=w_{f} \frac{S_{f}-u_{f}}{\delta_{f}}+w_{p} \frac{S_{p}-u_{p}}{\delta_{p}}
\end{equation}
where $u_{f}$, $\delta_{f}$, $u_{p}$, and $\delta_{p}$ denote the means and standard deviations of the inconsistency between appearance and motion feature and prediction error of all the normal training samples, respectively. $w_{f}$ and $w_{p}$ represent the weights of the two scores.

\section{Experimental results}
\label{sec:format}
\textbf{Implementation Details. }
We evaluate our approach on three public VAD benchmarks, including UCSD ped2 \cite{mahadevan2010anomaly}, CUHK Avenue \cite{lu2013abnormal}, and ShanghaiTech \cite{luo2017revisit} datasets. Following \cite{liu2021hybrid, yu2020cloze}, we train our model on the patches with foreground objects instead of the whole video frames. In advance, all foreground objects are extracted from original videos for the training and testing samples. RoI bounding boxes identify foreground objects. For each RoI, a spatial-temporal cube (STC) composed of the object in the current frame and the content in the same region of previous $t$ frames will be built, where the hyper-parameter $t$ is set to 4. And the width and height of STCs are resized to 32 pixels. The corresponding optical flows are generated by FlowNet2 \cite{ilg2017flownet}, and the STCs for optical flows are built in a similar way. Due to the existence of many objects in a frame, we select the maximum anomaly score of all objects as the anomaly score of a frame. We adopt Adam optimizer with an initial learning rate of $2e^{-4}$, decayed by 0.8 after every ten epochs. The batch size and epoch number of Ped2, Avenue, and ShanghaiTech are set to $(128, 60)$, $(128, 40)$, $(256, 40)$. $\lambda_{int}$, $\lambda_{gd}$, $\lambda_{sim}$, and $\lambda_{model}$ for Ped2, Avenue, and ShanghaiTech are set to $(1, 1, 1, 1)$, $(1, 1, 1, 1)$, $(1, 1, 10, 1)$. Then the error fusing weights $w_{f}$, $w_{p}$ for Ped2, Avenue, and ShanghaiTech are set to $(1, 0.01)$, $(0.2, 0.8), (0.4, 0.6)$.
\begin{table}
\footnotesize
    \begin{center}
	\caption{AUROC (\%) comparison between the proposed AMSRC-Net and state-of- the-art VAD methods on three public benchmarks.}
	\label{t1}
	\begin{tabular}{c|ccc}
    \hline
    Methods            &   UCSD Ped2 & CUHK Avenue & ShanghaiTech \\ \hline
    ConvAE\cite{hasan2016learning}      & 90            & 70.2          & N/A           \\
			 ConvLSTM-AE\cite{luo2017remembering} & 88.1          & 77            & N/A           \\
			 MemAE\cite{gong2019memorizing}       & 94.1          & 83.3          & 71.2          \\
			Frame-Pred.\cite{liu2018future}  & 95.4          & 85.1          & 72.8          \\
			 MNAD-R\cite{park2020learning}      & 97            & 88.5          & 70.5          \\
			 VEC\cite{yu2020cloze}         & 97.3          & 90.2          & 74.8          \\
			 AMC\cite{nguyen2019anomaly}         & 96.2          & 86.9          & N/A           \\
			 AnoPCN\cite{ye2019anopcn}      & 96.8          & 86.2          & 73.6          \\
			 AMMC-Net\cite{cai2021appearance}    & 96.6          & 86.6          & 73.7          \\
			 $\text{HF}^{2}$-VAD\cite{liu2021hybrid}     & 99.3          & 91.1          & 76.2          \\ \hline
    \textbf{AMSRC-Net}            & \textbf{99.5}      & \textbf{93.8}       & \textbf{76.6}        \\ \hline
	\end{tabular}
	\end{center}
\end{table}

\noindent
\textbf{Evaluation Criterion. }
We follow the widely popular evaluation metric in video anomaly detection \cite{liu2018future, liu2021hybrid, cai2021appearance} and evaluate our method using the frame-level area under the ROC curve (AUC) metric. The ROC curve is measured by varying the threshold over the anomaly score. Higher AUC values represent better performance for anomaly detection.

\begin{figure}[ht]
	\centering
	\includegraphics[width=8cm]{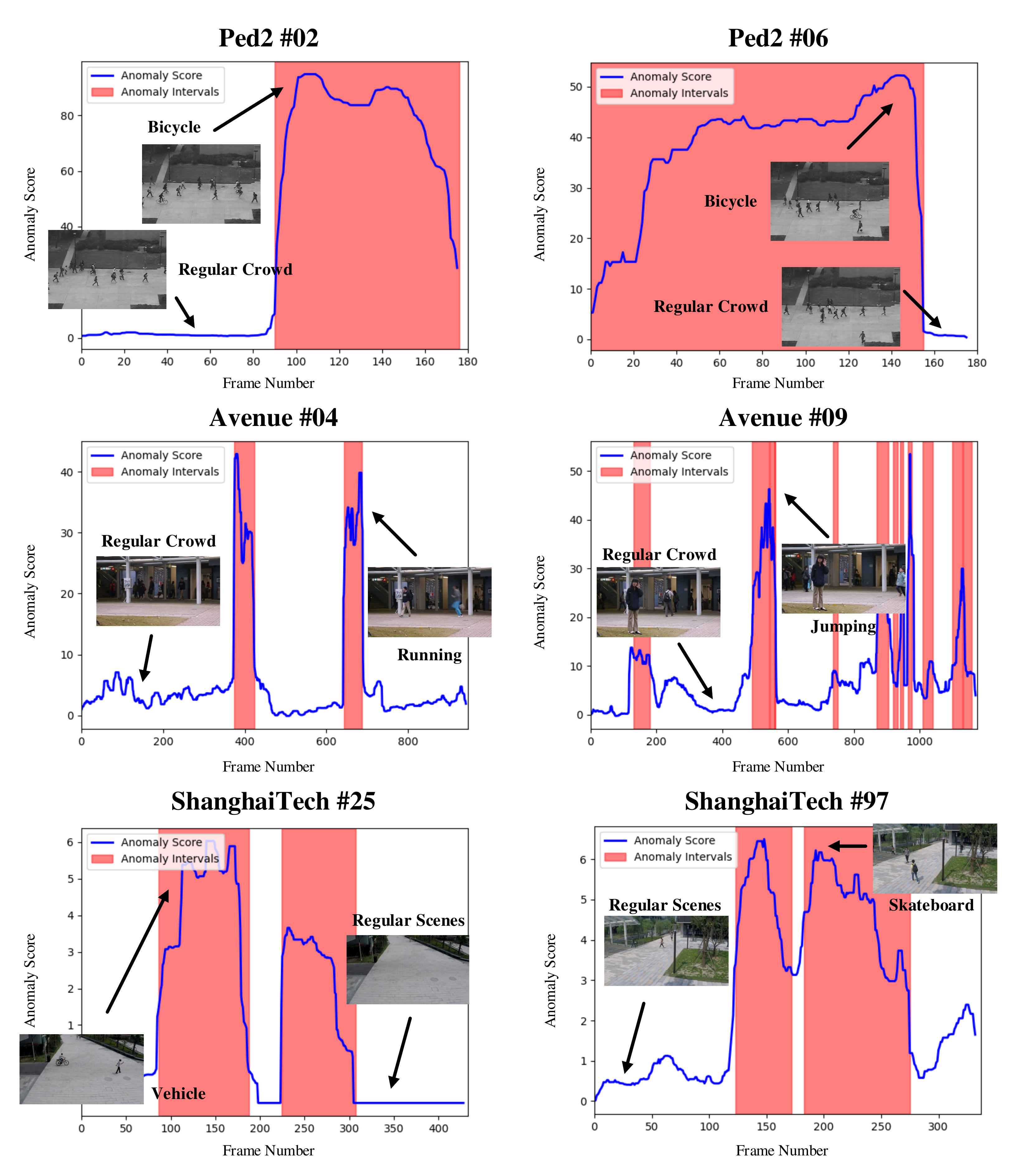}
	\caption{Anomaly score curves of some testing video clips. Red regions represent ground truth anomalous frames.}
	\label{p5}
\end{figure}

\noindent
\textbf{Anomaly Detection Results. }
To evaluate the performance of our AMSRC-Net, anomaly detection is performed on three public benchmarks. Examples in Figure \ref{p5} show anomaly score curves of some testing video clips. The anomaly score is calculated by Equation \ref{e5} and can be utilized to detect anomalies. The red regions denote the ground truth anomalous frames. As can be seen, the anomaly score of a video clip rises when anomalies occur, and decreases when anomalies disappear, which shows that our method can spot the anomalies accurately.

To our best knowledge, we compare our AMSRC-Net with several SOTA methods, and the results are summarized in Table \ref{t1}. It is evident that AMSRC-Net outperforms compared SOTA methods on all three benchmarks. In particular, we note that the proposed method achieves 93.8\% frame-level AUROC on CUHK Avenue, which is the best performance achieved on Avenue currently and exceeds the SOTA performance by 2.7\%.

\begin{table}
\footnotesize
\begin{center}
\caption{
Ablation studies of each component in our AMSRC-Net on the CUHK Avenue dataset.
}
\label{table:ablation}
\begin{tabular}{c|ccc|c}
\hline
         Index   & Optical Flow & Semantic Consistency & FGFM & AUC (\%)\\ \hline
  $\mathcal{A}$            & \XSolidBrush      & \XSolidBrush      & \XSolidBrush & 90.6      \\
 $\mathcal{B}$            & \CheckmarkBold      & \XSolidBrush           & \XSolidBrush & 90.8            \\
 $\mathcal{C}$            & \CheckmarkBold        & \XSolidBrush        & \CheckmarkBold & 91.2            \\
 $\mathcal{D}$           & \CheckmarkBold  &  \CheckmarkBold  & \XSolidBrush  & 92.5 \\
 $\mathcal{E}$   &  \CheckmarkBold  & \CheckmarkBold & \CheckmarkBold & 93.8 \\
\hline
\end{tabular}
\end{center}
\end{table}

\noindent
\textbf{Ablation Studies. }
We perform corresponding ablation studies to analyze the impact of different components of AMSRC-Net, including optical flow (motion) stream, consistency loss, and FGFM. The results are showed in Table \ref{table:ablation}. We can see that the introduction of optical flow brings a trivial improvement ($\mathcal{A}$ vs. $\mathcal{B}$). After the establishment of semantic consistency between appearance and motion, the performance is significantly enhanced ($\mathcal{B}$ vs. $\mathcal{D}$), showing the vital correlation between the two modalities for VAD is captured. Furthermore, the FGFM significantly enhances the AUC score by 1.3\% based on semantic consistency ($\mathcal{D}$ vs. $\mathcal{E}$), proving the effectiveness of our idea.

\begin{figure}[ht]
	\centering
	\includegraphics[width=8cm]{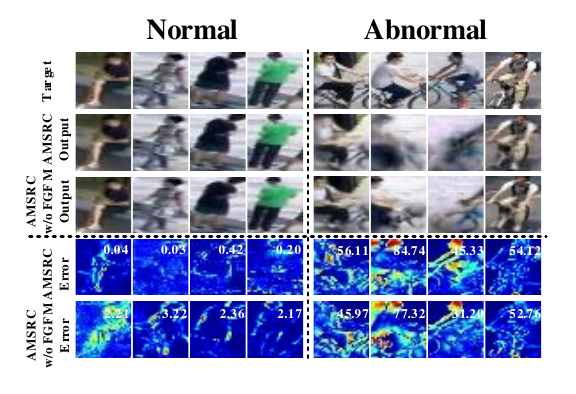}
	\caption{Visualization examples of the ground truth frames (Target), completed frames by AMSRC (AMSRC Output), completed frames by AMSRC without FGFM (AMSRC w/o FGFM Output), completion errors by AMSRC (AMSRC Error), and completion errors by AMSRC without FGFM (AMSRC w/o FGFM Error).}
	\label{p6}
\end{figure}

To show that our proposed FGFM can enlarge the gap in the quality of the predicted frame between normal and abnormal data, the visualized results of representative normal/abnormal events are demonstrated in Figure \ref{p6}. As we can see, the FGFM can help to produce larger differences for abnormal events, and these differences are observed in regions with motion behavior semantics. Such observations imply that AMSRC-Net pays more attention to high-level behavior semantics for anomalies.

\section{Conclusion}
\label{sec:pagestyle}

This paper presents a framework of Appearance-Motion Semantics Representation Consistency that uses the gap of appearance and motion semantic representation consistency between normal and abnormal data to detect anomalies. We design a two-stream encoder to extract normal samples' appearance and motion features and add constraints to strengthen their consistent semantics so that abnormal ones with lower consistency can be identified. Moreover, the lower consistency of appearance and motion features of anomalies can be fused by the flow-guided fusion module to affect the quality of predicted frames, making anomalies produce larger prediction differences. Experimental results on three public benchmarks show that our method performs better than state-of-the-art approaches.

\vfill\pagebreak

\bibliographystyle{IEEEbib}
\bibliography{strings,refs}

\end{document}